\documentclass[letterpaper]{article} 
\usepackage{aaai2026}  
\usepackage{times}  
\usepackage{helvet}  
\usepackage{courier}  
\usepackage[hyphens]{url}  
\usepackage{graphicx} 
\usepackage{natbib}  
\usepackage{caption} 
\usepackage{algorithm}
\usepackage{algpseudocode}
\usepackage[utf8]{inputenc} 
\usepackage[T1]{fontenc}    
\usepackage{booktabs}       
\usepackage{amsfonts}       
\usepackage{nicefrac}       
\usepackage{microtype}      
\usepackage{xcolor}         
\usepackage{graphicx}
\usepackage{algorithm}
\usepackage{verbatim}
\usepackage{amsmath}
\usepackage{multirow}
\usepackage{newfloat}
\usepackage{listings}
\DeclareCaptionStyle{ruled}{labelfont=normalfont,labelsep=colon,strut=off} 
\floatstyle{ruled}
\newfloat{listing}{tb}{lst}{}
\floatname{listing}{Listing}
\title{Learning to be Reproducible: Custom Loss Design for Robust Neural Networks}
\author{
    Waqas Ahmed\textsuperscript{\rm 1},
    Sheeba Samuel\textsuperscript{\rm 2},
    Kevin Coakley\textsuperscript{\rm 3},
    Birgitta Koenig-Ries\textsuperscript{\rm 1},
    Odd Erik Gundersen\textsuperscript{\rm 3}
}
\affiliations{
    \textsuperscript{\rm 1}Friedrich Schiller University Jena, Jena, Germany\\
    \textsuperscript{\rm 2}University of Technology Chemnitz, Chemnitz, Germany\\
    \textsuperscript{\rm 3}Norwegian University of Science and Technology, Trondheim, Norway\\
    
    waqas.ahmed@uni-jena.de, sheeba.samuel@informatik.tu-chemnitz.de, 
    kcoakley@sdsc.edu,
    birgitta.koenig-ries@uni-jena.de,
    odderik@ntnu.no
}

\begin{document}

\maketitle

\begin{abstract}
 To enhance the reproducibility and reliability of deep learning models, we address a critical gap in current training methodologies: the lack of mechanisms that ensure consistent and robust performance across runs. Our empirical analysis reveals that even under controlled initialization and training conditions, the accuracy of the model can exhibit significant variability. To address this issue, we propose a Custom Loss Function (CLF) that reduces the sensitivity of training outcomes to stochastic factors such as weight initialization and data shuffling. By fine-tuning its parameters, CLF explicitly balances predictive accuracy with training stability, leading to more consistent and reliable model performance. Extensive experiments across diverse architectures for both image classification and time series
forecasting demonstrate that our approach significantly improves
training robustness without sacrificing predictive performance. These results establish CLF as an effective and efficient strategy for developing more stable, reliable and trustworthy neural networks.
\end{abstract}

\section{Introduction}
\label{sec:intro}

Deep Learning (DL) models have become foundational across a wide range of applications, including healthcare diagnostics, autonomous systems, and financial forecasting, due to their remarkable ability to learn complex representations from large-scale data. Despite this success, achieving consistent and trustworthy performance from these models remains a significant and under-addressed challenge. Even when training conditions such as architecture, hyperparameters, and datasets are fixed, models often yield substantially different results across runs. This variability arises from algorithmic sources of randomness such as weight initialization, data shuffling, and optimizer behavior, which affect the trajectory of model training and lead to inconsistent outcomes.
Recent studies have highlighted the sensitivity of deep neural networks to such stochastic factors, revealing that even minor changes in initialization can cause large deviations in final model performance \citep{summers2021nondeterminism}. A common practice for controlling stochastic effects in deep learning is to fix the random seed during training. This does not directly reduce the influence of stochastic factors; instead, it determines the sequence of all random operations through pseudo-random number generators (PRNGs), ensuring that the same sequence is reproduced in every run with that seed. As a result, repeating an experiment with the same seed yields identical outcomes. However, when a different seed is used, the sequence of random operations changes, which in turn alters the training trajectory and can lead to substantially different results. Consequently, model performance remains sensitive to the choice of seed, and variability introduced by stochastic factors persists. While seed fixing enables a narrow form of reproducibility for a specific experimental setup, it does not ensure robustness in the broader sense needed for reliable conclusions across different runs \citep{pham2020problems}. Secondly, another common approach is to report averaged metrics over multiple runs with different seeds. This provides a more comprehensive view of model performance by sampling across multiple PRNG sequences, thereby capturing a broader range of variability. While statistically more sound than relying on a single seed, this approach comes at a substantial computational cost. In many cases, it requires 25 or more complete training runs to obtain stable estimates \citep{renard2020variability,bouthillier2019unreproducible}, making it impractical for large-scale experiments, resource-constrained environments, or real-time development settings.

Our research addresses this limitation by proposing a method that reduces training variability directly within the learning objective. Drawing inspiration from the need for more trustworthy and stable deep learning systems, we introduce a Custom Loss Function (CLF) designed to regularize the training process by penalizing fluctuations in prediction confidence and output loss. Instead of removing randomness altogether, CLF mitigates its downstream effects, leading to more stable convergence and reduced run-to-run variability. This approach is lightweight and compatible with existing architectures and training pipelines.
In developing this method, we conducted a systematic investigation into the algorithmic sources of variability, employing fixed-identical training conditions to isolate and measure the impact of random components. Additionally, we explored the sensitivity of training to hyperparameters within the loss function itself, uncovering the importance of tuning both the weight of the variability penalty and the timing of its integration. In particular, we find that applying CLF earlier and maintaining it throughout training strikes a better balance between stability and learning efficiency.

We validate our approach through extensive experiments on image classification tasks using CIFAR-10 and CIFAR-100 \citep{krizhevsky2009learning} with architectures including ResNet \citep{he2016deep}, VGG \citep{DBLP:journals/corr/SimonyanZ14a}  , and ShuffleNet \citep{DBLP:conf/eccv/MaZZS18}. We also demonstrate generalizability by applying CLF to time series forecasting models (Autoformer~\citep{10.5555/3540261.3541978}, NLinear~\citep{10.1609/aaai.v37i9.26317}, and iTransformer~\citep{liu2023itransformer}) on the ETTh1~\citep{haoyietal-informer-2021} dataset. Results consistently show that CLF reduces standard deviation in model performance, sometimes by over 70\%, without compromising accuracy.
Our key contributions are:
 \begin{enumerate}
 \item \textbf{Custom Loss Function}: We propose a new loss formulation that explicitly incorporates variability control, significantly reducing performance fluctuations between training runs.
 \item \textbf{Duration-Sensitive Effectiveness}: We analyze the impact of CLF activation duration and show that longer exposure during training consistently leads to better performance and stability, while late-stage activation offers limited benefit.

 \item \textbf{Cross-Domain Generalizability}: We demonstrate that CLF improves stability in both image and time series domains, confirming its broad applicability across architectures and tasks.
 \end{enumerate}
Together, these findings pave the way for more reliable and trustworthy deep learning systems by addressing variability not only at the evaluation level, but within the training dynamics themselves.

\section{Related work}
\label{sec:relatedwork}

Technical robustness is a critical component of trustworthy artificial intelligence, particularly in systems deployed in dynamic or high-stakes environments. Kaur et al.~\cite{10.1145/3491209} provide a comprehensive survey, emphasizing on the need for systems that are resilient to perturbations and implementation variability in order to ensure reliable behavior. Gundersen et al. \cite{gundersen2023reporting} focus specifically on robustness against algorithmic randomness in neural network training. They demonstrate that stochastic elements such as weight initialization, data shuffling, and non-deterministic operations can introduce significant variability in model performance. Their findings show that this variability is often underestimated. They propose methodological standards including at least 25 repeated training runs to support statistically sound conclusions. Importantly, they argue that robustness to randomness is not a secondary technical detail but a prerequisite for drawing trustworthy scientific and empirical claims. This framing aligns with the EU’s High-Level Expert Group on AI, which defines trustworthiness through principles such as robustness, accountability, and transparency. In deep learning, achieving robustness to randomness is therefore not only a matter of mitigating noise but a foundation for consistent and reliable behavior in real-world deployment.

Building on these conceptual foundations, recent empirical work has highlighted how the lack of robustness to randomness manifests in practice. Bouthillier et al.~\cite{bouthillier2019unreproducible} present a critical assessment of reproducibility failures in machine learning and argue that many of these issues stem from insufficient control over experimental variability. Their study illustrates how minor implementation details, random seed choices, or system-level factors can lead to substantial performance fluctuations. Zhuang et al.~\cite{zhuang2022randomness} extend this line of inquiry by examining how tool-level randomness, such as that introduced by software libraries, compilers, and system-level abstractions, can influence the trajectory of neural network training. Their results demonstrate that seemingly identical configurations can produce divergent models because of low-level sources of nondeterminism.
 Similarly, Summers et al.~\cite{summers2021nondeterminism} focus on the instability introduced by optimization procedures such as stochastic gradient descent. They show that model outcomes can vary significantly across runs even when initialization, data, and hyperparameters are held constant, pointing to a deeper technical instability in DL optimization dynamics. Ahmed et al.~\cite{ahmed2022managing} propose practical strategies to manage pseudo-randomness, including consistent seeding and systematic logging of random state. They frame this as essential for improving both trustworthiness and experimental reliability. However, as Summers et al.~\cite{summers2021nondeterminism} show, such control measures alone do not eliminate instability; deeper algorithmic sensitivity remains a challenge. Yi et al.~\cite{ji2023randomness} reinforce this point by analyzing the effect of randomness on evaluation metrics. They recommend multi-run reporting and controlled seed strategies, yet acknowledge that determining a sufficient number of runs remains unresolved. Picard~\cite{picard2021torch} provides empirical evidence on how random seed selection can dramatically affect reported results in computer vision models.

In response to these ongoing challenges, our work shifts focus from external mitigation strategies to an internal algorithmic solution. We introduce a CLF that explicitly regularizes the training dynamics to reduce variance in model outcomes. Rather than relying on repeated training or strict seed control, CLF stabilizes the learning trajectory itself. This approach enhances technical robustness to randomness and supports the development of neural networks that behave consistently under varying stochastic conditions.

\begin{figure*}[t]
\centering
\includegraphics[width=1\textwidth]{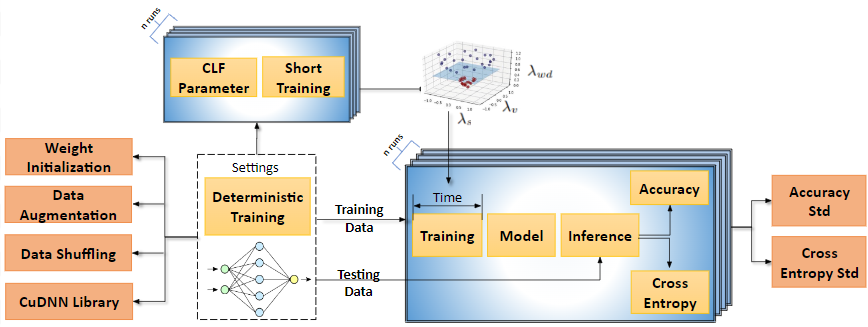} 
\vspace{0.5em}
\caption{Methodological overview (classification task): This diagram illustrates the process of fixed identical training conducted twenty times (n=20) to evaluate the efficacy of  CLF on different set of dataset and models. We performed deterministic training by controlling for random sources. Model variability is assessed in terms of the standard deviation of accuracy and cross-entropy loss, aiming to quantify the robustness of the training process. }
\vspace{-1em}
\label{fig:methodology}
\end{figure*}

\section{Custom loss function}
\label{sec:vml}

The proposed custom loss function (CLF) is designed to mitigate the stochastic behavior commonly observed in deep neural network training, including variability in model performance across different runs due to random initialization, data shuffling, and system-level nondeterminism.  
CLF improves the consistency of optimization by enhancing both the stability of gradient updates and the coherence of predictions within and across mini-batches.  

It consists of three components:  
\textbf{(1) Cross-Entropy Loss (CEL)} (Section~\ref{sec:cel}),  
\textbf{(2) Stable Loss (SL)} (Section~\ref{sec:sl}), and  
\textbf{(3) Variance Penalty Loss (VPL)} (Section~\ref{sec:vpl}).  

The total loss is defined as:

\begin{equation}
\begin{split}
L(\theta; x, y)
&= \mathrm{CEL}(\theta; x, y) \\
&\quad + \lambda_s\, \mathrm{SL}(\theta; x, y) \\
&\quad + \lambda_v\, \mathrm{VPL}(\theta; x, y)
\end{split}
\label{eq:vml_total}
\end{equation}

where $\theta$ are the model parameters, $(x, y)$ are the input samples and ground-truth labels, and $\lambda_s$, $\lambda_v$ are scalar hyperparameters controlling the influence of SL and VPL.

Each term plays a distinct role in controlling instability and improving robustness to randomness.

\subsection{Cross-Entropy Loss (CEL)}
\label{sec:cel}

The Cross-Entropy Loss is the standard objective for classification tasks, measuring the dissimilarity between the predicted class distribution and the true label distribution:
\begin{equation}\label{eq:CEL}
\mathrm{CEL}(\theta; x, y) = -\frac{1}{N} \sum_{i=1}^{N} \log p(y_i | x_i; \theta),
\end{equation}
where $p(y_i | x_i; \theta)$ is the predicted probability of the correct class $y_i$ given input $x_i$, and $N$ is the batch size.

While effective for guiding the model toward correct predictions, CEL alone does not impose any constraint on the stability of the optimization trajectory.  
In settings with noisy labels or class imbalance, small prediction errors can result in low confidence scores and correspondingly large-magnitude gradients.  
This sensitivity is reflected in its gradient:
\begin{equation}
\nabla_{\theta} \mathrm{CEL} = - \frac{1}{N} \sum_{i=1}^{N} \frac{1}{p(y_i | x_i; \theta)} \nabla_{\theta} p(y_i | x_i; \theta),
\end{equation}
which can grow rapidly when $p(y_i | x_i; \theta)$ is small, causing unstable updates.

To counteract this, CLF incorporates two additional terms:  
SL focuses on inter-epoch stability, and VPL enforces intra-batch, per-class consistency.

\subsection{Stable Loss (SL)}
\label{sec:sl}

SL targets temporal stability during training by penalizing sudden changes in the loss value between consecutive epochs.  
It is defined as:
\begin{equation}\label{eq:SL}
\mathrm{SL}(\theta; x, y) = \left| \mathrm{CEL}(\theta; x, y) - \mathrm{CEL}_{\mathrm{prev}} \right|,
\end{equation}
where $\mathrm{CEL}_{\mathrm{prev}}$ is the cross-entropy loss from the previous epoch.

Its gradient is:
\begin{equation}
\nabla_{\theta} \mathrm{SL} = \mathrm{sign}\!\left(\mathrm{CEL} - \mathrm{CEL}_{\mathrm{prev}}\right) \cdot \nabla_{\theta} \mathrm{CEL},
\end{equation}
where the sign term controls the penalty direction.  
Unlike traditional regularizers that act on parameters, SL operates at the loss level, making it architecture-agnostic.

\subsection{Variance penalty loss (VPL)}
\label{sec:vpl}

While SL regulates the temporal aspect of training, VPL focuses on spatial consistency within a mini-batch. It penalizes variance in model predictions across samples belonging to the same class, thereby promoting robustness and discouraging the model from overfitting to batch-specific noise. The VPL is defined as an average over per-class variances within the mini-batch:

\begin{equation}\label{eq:VPL}
\mathrm{VPL}(\theta; x, y) = \frac{1}{|\mathcal{C}|} \sum_{j \in \mathcal{C}} \mathrm{Var}_j,
\end{equation}

where $\mathcal{C}$ is the set of class labels present in the mini-batch, and $\mathrm{Var}_j$ is the variance for class $j$, computed as:

\begin{equation}\label{eq:Varj}
\mathrm{Var}_j = \frac{1}{m_j} \sum_{i \in S_j} \left( f_j(x_i; \theta) - \bar{f}_j \right)^2,
\end{equation}

with $S_j = \{ i \mid y_i = j \}$ denoting the set of samples with true label $j$, $m_j = |S_j|$, and the class-mean logit $\bar{f}_j$ given by:

\begin{equation}\label{eq:meanlogit}
\bar{f}_j = \frac{1}{m_j} \sum_{i \in S_j} f_j(x_i; \theta).
\end{equation}

\paragraph{Gradient of VPL.}  
The gradient of VPL with respect to $\theta$ is:

\begin{equation}\label{eq:gradVPL}
\nabla_\theta \mathrm{VPL}(\theta; x, y) = \frac{1}{|\mathcal{C}|} \sum_{j \in \mathcal{C}} \nabla_\theta \mathrm{Var}_j,
\end{equation}

where:

\begin{equation}
\begin{split}
\nabla_\theta \mathrm{Var}_j
&= \frac{2}{m_j} \sum_{i \in S_j}
   \left( f_j(x_i; \theta) - \bar{f}_j \right) \\
&\quad \times
   \left(
       \nabla_\theta f_j(x_i)
       - \frac{1}{m_j} \sum_{k \in S_j} \nabla_\theta f_j(x_k)
   \right)
\end{split}
\label{eq:gradVarj_full}
\end{equation}

Since
\begin{equation}
\sum_{i \in S_j} \left( f_j(x_i; \theta) - \bar{f}_j \right) = 0,
\end{equation}
the second term in \eqref{eq:gradVarj_full} cancels out, simplifying the gradient to:

\begin{equation}\label{eq:gradVarj_simplified}
\nabla_\theta \mathrm{Var}_j = \frac{2}{m_j} \sum_{i \in S_j} \left( f_j(x_i; \theta) - \bar{f}_j \right) \nabla_\theta f_j(x_i).
\end{equation}

Finally, the gradient of VPL becomes:

\begin{equation}
\begin{split}
\nabla_\theta \mathrm{VPL}(\theta; x, y)
&= \frac{2}{|\mathcal{C}|} \sum_{j \in \mathcal{C}} \sum_{i \in S_j}
    \frac{1}{m_j} \left( f_j(x_i; \theta) - \bar{f}_j \right) \\
&\quad \times \nabla_\theta f_j(x_i)
\end{split}
\label{eq:finalGradVPL}
\end{equation}

Minimizing this term encourages tighter clustering of same-class logits in the output space, thereby reducing intra-class variability, mitigating the effects of stochastic training factors, and improving run-to-run robustness.

\section{Methodology}
\label{sec:methodology}

 To assess the effectiveness of CLF, we developed a systematic experimental protocol that isolates stochastic effects and quantifies variance in both performance and training behavior. Figure~\ref{fig:methodology} outlines our approach.

We began by establishing deterministic baselines through strict control of known randomness sources. To do so, we fixed random seeds across all relevant libraries and disabled nondeterministic operations at the framework level, such as cuDNN benchmarking and parallel kernel execution. This ensured that observed variability arises only from inherent stochastic effects not eliminated by seeding. All experiments were conducted under identical hardware, software, and hyperparameter settings. To evaluate sensitivity to initialization, we trained each model across 20 different seeds \( S = \{1, 2, \dots, 20\} \). Each training run used the same architecture and optimization configuration, enabling a controlled analysis of run-to-run variability. We adopted a from-scratch training protocol, following the recommendations of Summers et al.~\cite{summers2021nondeterminism}, in which each model is trained independently from randomized initial weights. This approach captures the full variability introduced by stochastic components in model initialization and optimization, and avoids bias from warm-started models or transfer learning. Our experiments included both image classification and time series forecasting tasks. We quantified performance variability across seeds by measuring the standard deviation of test accuracy for classification tasks and the standard deviation of Mean Absolute Error (MAE) for time series forecasting.

CLF hyperparameters (\( \lambda_s \) and \( \lambda_v \)) were optimized using the Optuna framework~\cite{10.1145/3292500.3330701}. The tuning objective focused on minimizing validation variance rather than maximizing raw performance, in line with our goal of reducing outcome instability. Optuna allowed for efficient exploration of the parameter space without relying on exhaustive grid-based evaluation. Overall, our methodology is designed to assess whether CLF can improve the determinism and reliability of deep learning training, particularly in settings where significant variability persists even under tightly controlled experimental conditions.

\subsection{Hyperparameter optimization with Optuna:}
\label{sec:optuna}

We optimized the CLF parameters using Optuna, a hyperparameter optimization framework that efficiently explores the search space through a Bayesian sampling approach. Unlike grid search, which exhaustively evaluates all combinations, Optuna dynamically prioritizes promising configurations, reducing computational overhead. Algorithm~\ref{alg:optuna} outlines the tuning process, focusing on the key parameters: The variance penalty weight (\(\lambda_v\)) controls output variance reduction, the stability weight (\(\lambda_s\)) regulates stable loss contribution, and the variance penalty weight decay (\(\lambda_{\text{wd}}\)) prevents over-penalization. This algorithm identifies promising configurations that can be utilized in our experiments for further analysis.

\begin{algorithm}[h]
\caption{Optuna hyperparameter optimization for custom loss function}
\label{alg:optuna}
\begin{algorithmic}[1]
\Require Search range for $\lambda_v$, $\lambda_s$, $\lambda_{\text{wd}}$
\State \textbf{Initialize} Optuna study with minimization objective
\For{each trial in range(num\_trials)}
    \State Sample $\lambda_v, \lambda_s, \lambda_{\text{wd}}$ using log-uniform distribution
    \State Initialize model with sampled parameters
    \State Optimize model using SGD and learning rate scheduler
    \State Evaluate test accuracy across multiple seeds
    \State Compute normalized accuracy and standard deviation
    \State Compute objective score: $\text{score} = \text{norm\_std} - \text{norm\_acc}$
    \If{score < best\_score}
        \State Update $best\_params \gets \{\lambda_v, \lambda_s, \lambda_{\text{wd}}, \text{score}\}$
    \EndIf
\EndFor
\State \textbf{Return} $best\_params$
\end{algorithmic}
\end{algorithm}

\subsection{Experimental setup}
\label{sec:setup}

Our experiments are conducted across two domains: image classification and time series forecasting. For the image classification tasks, we utilize the CIFAR-10 and CIFAR-100 datasets as shown in Table \ref{tab:datasets}. We test three widely used convolutional neural network architectures: ResNet, VGG-16, and ShuffleNet-V2, each selected for its proven effectiveness in classification tasks. Training follows a cosine decay learning rate schedule with an initial peak of 0.40, a batch size of 512, momentum set to 0.9, and a weight decay of $5 \times 10^{-4}$. The CLF parameters (\( \lambda_s \) and \( \lambda_v \)) are tuned separately for each model-dataset pair. All training runs are performed from scratch using 20 distinct random seeds to evaluate robustness under initialization variability. The primary evaluation metric is the standard deviation of test accuracy across these runs. All experiments are conducted using PyTorch~\cite{paszke2019pytorch} on a compute environment with 64 CPU cores, 512 GB of RAM, and NVIDIA A100 GPU.

For time series forecasting, we employ the ETTh1 dataset, a subset of the Electricity Transformer Temperature (ETT) dataset. This dataset contains two years of hourly-level measurements of transformer oil temperature and six related power load features, collected from two counties in China. The dataset is split into training, validation, and test sets using a 12:4:4 month ratio. The forecasting task involves predicting future oil temperature values based on historical multivariate input sequences. We use three neural architectures tailored for long-sequence forecasting: NLinear, Autoformer, and iTransformer. For each model, we adopt the training configurations and hyperparameters as specified in the respective original publications. This experimental design enables a direct comparison of CLF’s performance across different modalities and model types, allowing us to assess its contribution to robustness under a broad range of conditions.

\begin{table}[h]
\vspace{0.5em}
\caption{Datasets, networks, and training settings.}
\vspace{0.5em}
\centering
\resizebox{0.45\textwidth}{!}{
\begin{tabular}{lll} 
\toprule
Dataset    & Train/ Validation/Test split      & Network  \\
\midrule
CIFAR-10   & 50,000 /- /10,000       & ResNet-14  \\
           &                       & VGG-16     \\
           &                       & ShuffleNet-V2 \\
\midrule
CIFAR-100  & 50,000 /- /10,000       & ResNet-32  \\
           &                       & VGG-16     \\
           &                       & ShuffleNet-V2 \\
\midrule
ETTh1      & 12/4/4 Months         & NLinear    \\
           &                       & Autoformer \\
           &                       & iTransformer \\
\bottomrule
\end{tabular}}
\label{tab:datasets}
\end{table}

\section{Results and discussion}
\label{sec:results}

This section presents a comprehensive evaluation of the CLF across both image classification and time series forecasting tasks.  Overall, we executed 280 fixed, identical training sessions for the classification task, along with a few hyperparameter searches, totaling approximately 230 hours of GPU time. For the time series experiments, we conducted 400 fixed identical runs, requiring more than 250 hours of GPU time.

\subsection{Impact of CLF on image classification}

\begin{table*}[t]
\centering
\vspace{0.5em}
\caption{Reduction of variability through the introduction of CLF.}
\vspace{1em}
\label{tab:variability}
\resizebox{1\textwidth}{!}{ 
\begin{tabular}{@{}llccccc@{}} 
\toprule
Training & Dataset & Model & Mean Acc. (\%) & \begin{tabular}[c]{@{}c@{}}Acc. SD \\ (\%)\end{tabular} & \begin{tabular}[c]{@{}c@{}}Cross-Ent. SD \\ (\%)\end{tabular} & \begin{tabular}[c]{@{}c@{}} Avg. Var. \\ Reduction (\%)\end{tabular} \\
\midrule
\multirow{6}{*}{Without CLF} 
 & \multirow{3}{*}{CIFAR-10}  & ResNet-14       & 88.2 & 0.14 $\pm$ 0.04 & 0.008 $\pm$ 0.002 & -  \\
 &                             & VGG-16          & 93.8 & 0.12 $\pm$ 0.04 & 0.007 $\pm$ 0.002 & -  \\
 &                             & ShuffleNet-V2   & 90.8 & 0.20 $\pm$ 0.07 & 0.008 $\pm$ 0.002 & -  \\
\cmidrule(lr){2-7}
 & \multirow{3}{*}{CIFAR-100} & ResNet-14       & 62.6 & 0.16 $\pm$ 0.05 & 0.01 $\pm$ 0.004  & -  \\
 &                             & VGG-16          & 74.0 & 0.17 $\pm$ 0.06 & 0.008 $\pm$ 0.002 & -  \\
 &                             & ShuffleNet-V2   & 67.9 & 0.30 $\pm$ 0.06 & 0.017 $\pm$ 0.007 & -  \\
\midrule
\multirow{6}{*}{With CLF} 
 & \multirow{3}{*}{CIFAR-10}  & ResNet-14       & 88.3 & 0.08 $\pm$ 0.02 & 0.006 $\pm$ 0.001  & 42.2  \\
 &                             & VGG-16          & 93.7 & 0.11 $\pm$ 0.03 & 0.006 $\pm$ 0.002  & 12.2  \\
 &                             & ShuffleNet-V2   & 90.7 & 0.13 $\pm$ 0.04 & 0.003 $\pm$ 0.0008 & 33.9  \\
\cmidrule(lr){2-7}
 & \multirow{3}{*}{CIFAR-100} & ResNet-14       & 62.1 & 0.10 $\pm$ 0.04 & 0.02 $\pm$ 0.007  & 39.4  \\
 &                             & VGG-16          & 74.2 & 0.14 $\pm$ 0.03 & 0.006 $\pm$ 0.002  & 26.1  \\
 &                             & ShuffleNet-V2   & 67.8 & 0.07 $\pm$ 0.02 & 0.020 $\pm$ 0.008  & 77.1  \\
\bottomrule
\vspace{-1em}
\end{tabular}}
\end{table*}
 As shown in Table~\ref{tab:variability}, CLF consistently reduced variability across settings, although the extent of improvement varied by model and dataset. For ResNet-14 on CIFAR-10, CLF reduced the accuracy standard deviation from $0.14 \pm 0.04$ to $0.08 \pm 0.02$, corresponding to a 42.2\% reduction in variability. The calculation of this reduction incorporates both lower and upper error bounds to ensure that even minimal fluctuations are considered. Specifically, the expected variability is estimated as the average of the extremes:

\begin{equation}
\small
\text{Avg. Var. Reduction} = \frac{\text{Upper Bound Var.} + \text{Lower Bound Var.}}{2}
\end{equation}

This approach ensures a more robust and inclusive measure of training variability. Similar trends are observed in other configurations. ShuffleNet-V2 on CIFAR-10 yielded a 33.9\% reduction, while VGG-16 showed a smaller reduction of 12.2\%. The limited impact of CLF in the VGG-16 case is attributed to the already low baseline variability in the non-CLF setting, with a reported standard deviation of $0.12 \pm 0.04$. Since the variance was minimal to begin with, the opportunity for further reduction was inherently constrained. On CIFAR-100, ResNet-14 achieved a reduction of 39.4\%, VGG-16 improved by 26.1\%, and ShuffleNet-V2 achieved the most substantial reduction of 77.1\%. The particularly large reduction observed with ShuffleNet-V2 may be attributed to its lightweight architecture, which tends to be more sensitive to random seed variations and training instability. Our method appears especially beneficial in such cases, where even small improvements in stability can translate into substantial performance consistency gains. These results confirm that CLF is particularly effective in training environments with high initial variance or unstable optimization trajectories. By reducing sensitivity to stochastic factors, CLF produces models that maintain accuracy while exhibiting consistent behavior across repeated runs, enhancing training trustworthiness.

\subsection{Sample-based evaluation of CLF robustness}
\begin{table}[t]
\centering
\vspace{0.5em}
\caption{Comparison of ResNet-14 with and without CLF on CIFAR-10 across group sizes.}
\vspace{0.5em}
\label{tab:resnet14_cifar10}
\resizebox{1\columnwidth}{!}{
\begin{tabular}{@{}cccccc@{}}
\toprule
Group Size & \begin{tabular}[c]{@{}c@{}}Mean Acc.\\ (CLF)\end{tabular} & \begin{tabular}[c]{@{}c@{}}Std Dev\\ (CLF)\end{tabular} & \begin{tabular}[c]{@{}c@{}}Mean Acc.\\ (No CLF)\end{tabular} & \begin{tabular}[c]{@{}c@{}}Std Dev\\ (No CLF)\end{tabular} & \begin{tabular}[c]{@{}c@{}}Lower Std\\ Dev Group\end{tabular} \\
\midrule
5  & 88.29478 & 0.1572 & 88.26957 & 0.1848 & With CLF \\
10 & 88.29826 & 0.1827 & 88.26784 & 0.2024 & With CLF \\
15 & 88.29689 & 0.1890 & 88.26738 & 0.2074 & With CLF \\
\bottomrule
\end{tabular}}
\end{table}

To assess whether the robustness introduced by CLF generalizes beyond specific seed groupings, we conducted a controlled experiment using a fixed pool of 20 trained models. We randomly sampled 1000 subsets of size 5, 10, and 15 to simulate practical scenarios where only a limited number of models might be deployed. For each group, we calculated the mean accuracy and standard deviation separately for models trained with and without CLF. The key metric was the frequency with which one configuration achieved a lower standard deviation. As shown in Table~\ref{tab:resnet14_cifar10}, CLF resulted in more stable performance in most cases. For instance, in the group size of 15, CLF achieved a lower standard deviation in 75\% of the samples. This trend was consistent across all group sizes, showing that CLF improves performance consistency both at the individual model level and when models are evaluated in small ensembles. These results support the broader applicability of CLF in deployment settings where only a limited subset of trained models is available.

\subsection{CLF under overfitting conditions}
To investigate the behavior of our CLF under overfitting scenarios, we conducted experiments using the ResNet-32 architecture on the CIFAR-100 dataset. This setting, characterized by a relatively high-capacity model and a complex, fine-grained classification task, was chosen to examine how CLF performs when the model is likely to overfit. Unlike the improvements observed in smaller architectures such as ResNet-14, the results in this case revealed that CLF was ineffective in reducing variability. Across all evaluated group sizes (5, 10, and 15), models trained with CLF consistently showed higher standard deviation in accuracy compared to their non-CLF counterparts as shown in Table \ref{tab:resnet32_cifar100}. This indicates that under overfitting conditions, the variance regularization introduced by CLF may conflict with the model’s inherent learning dynamics, thereby amplifying rather than suppressing variability. However, it is important to note that the average accuracy remained stable, suggesting that CLF did not negatively impact overall classification performance. These findings suggest that while CLF is effective in controlling variability in appropriately scaled models, it does not offer the same benefit when applied to overparameterized settings where overfitting dominates.

\begin{table}[t]
\centering
\vspace{0.5em}
\caption{Comparison of ResNet-32 with and without CLF on CIFAR-100 across group sizes.}
\vspace{0.5em}
\label{tab:resnet32_cifar100}
\resizebox{1\columnwidth}{!}{
\begin{tabular}{@{}cccccc@{}}
\toprule
Group Size & \begin{tabular}[c]{@{}c@{}}Mean Acc.\\ (CLF)\end{tabular} & \begin{tabular}[c]{@{}c@{}}Std Dev\\ (CLF)\end{tabular} & \begin{tabular}[c]{@{}c@{}}Mean Acc.\\ (No CLF)\end{tabular} & \begin{tabular}[c]{@{}c@{}}Std Dev\\ (No CLF)\end{tabular} & \begin{tabular}[c]{@{}c@{}}Lower Std\\ Dev Group\end{tabular} \\
\midrule
5  & 62.3355 & 0.2154 & 62.5287 & 0.1920 & Without CLF \\
10 & 62.3356 & 0.2381 & 62.5300 & 0.2084 & Without CLF \\
15 & 62.3341 & 0.2455 & 62.5295 & 0.2133 & Without CLF \\
\bottomrule
\end{tabular}}
\end{table}

\subsection{Impact of CLF activation duration on training performance}
\label{sec:clf_timing}

\begin{figure}[ht]
\centering
\includegraphics[width=1\columnwidth]{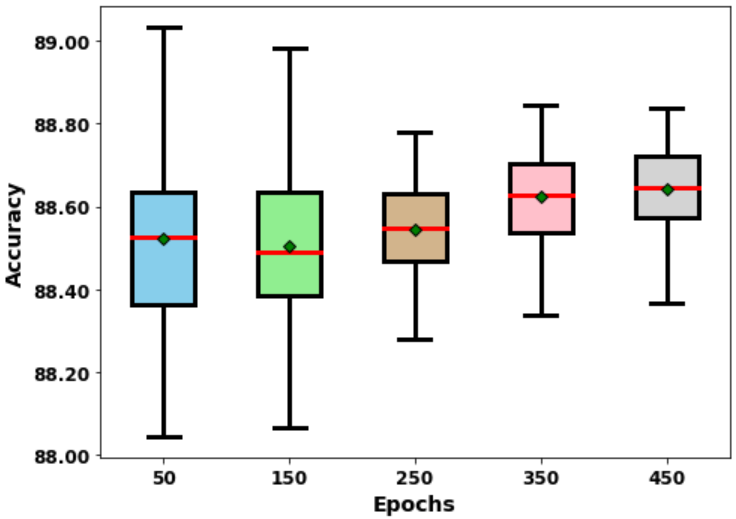}
\vspace{0.5em}
\caption{ Test accuracy distribution across 5 runs for different CLF activation durations on CIFAR-10 with ResNet-14. The x-axis indicates the total number of epochs during which the CLF was active out of a fixed 500-epoch training schedule.}
\vspace{0.5em}
\label{fig:box_plot}
\end{figure}

We investigated how the duration of CLF application during training affects model performance and variability. Since CLF combines cross-entropy with additional regularization, the key question is how long it should remain active within the training schedule. To assess this, we trained ResNet-14 on CIFAR-10 for 500 epochs, activating CLF during the final 50, 150, 250, 350, and 450 epochs, while using standard cross-entropy in the earlier stages. Each configuration was repeated with 5 random seeds to evaluate variability. The extended training schedule ensured full convergence and allowed a fair assessment of CLF's impact. Results (Figure~\ref{fig:box_plot}) indicate that longer CLF exposure improves both accuracy and stability. When CLF is applied for a significant portion of training, the model benefits from the stabilizing effects of Stable Loss (SL) and Variance Penalty Loss (VPL). In contrast, short-duration CLF usage, limited to the end of training, has minimal impact as the model has already converged based on cross-entropy, leaving little room for adjustment.

\subsection{Time series evaluation with CLF}
\begin{table*}[t]
\centering
\vspace{0.5em}
\caption{Reduction of prediction variability (SD of MAE) in time series forecasting using CLF.}
\vspace{1em}
\label{tab:timeseries_variability}
\resizebox{0.8\textwidth}{!}{
\begin{tabular}{@{}lccccl@{}}
\toprule
Model & Dataset & Horizon & MAE SD (Without CLF) & MAE SD (CLF) & Reduction (\%) \\
\midrule
Autoformer   & ETTh1 & 192 & 0.0151 & 0.0150 & 0.66 \\
NLinear      & ETTh1 & 192 & 0.0023 & 0.0021 & 8.70 \\
iTransformer & ETTh1 & 192 & 0.0015 & 0.0018 & $-20.00$ \\
Autoformer   & ETTh1 & 336 & 0.0146 & 0.0137 & 6.16 \\
NLinear      & ETTh1 & 336 & 0.0018 & 0.0013 & 27.78 \\
iTransformer & ETTh1 & 336 & 0.0025 & 0.0026 & –4.00 \\
NLinear      & ETTh1 & 720 & 0.0004 & 0.0003 & 25.00 \\
iTransformer & ETTh1 & 720 & 0.0044 & 0.0040 & 9.09 \\
\bottomrule
\end{tabular}}
\end{table*}

We evaluated the effectiveness of CLF in time series forecasting using the ETTh1 dataset with three models: Autoformer, NLinear, and iTransformer, tested across prediction horizons of 192, 336, and 720. Since the primary loss function for time series forecasting is Mean Squared Error (MSE), we combined CLF’s Stable Loss and Variance Penalization Loss (VPL) with MSE to assess variability reduction. As shown in Table~\ref{tab:timeseries_variability}, CLF noticeably reduced prediction variability, measured by the standard deviation of the Mean Absolute Error (MAE), in most cases. NLinear showed the most significant improvements, with variability reductions of 8.70\%, 27.78\%, and 25.00\% at horizons 192, 336, and 720, respectively. This is likely due to NLinear’s simple, structure-free design, which lacks stabilization mechanisms, making it more susceptible to stochastic variation and more responsive to CLF. The benefit of CLF increased with longer prediction horizons, where greater uncertainty amplified instability. For instance, Autoformer reduced MAE variability by only 0.66\% at horizon 192 but by 6.16\% at horizon 336. In contrast, iTransformer was less affected by CLF. At horizons 192 and 336, variability slightly increased, with a modest 9.09\% reduction at horizon 720. This limited impact likely stems from iTransformer's inherent stability, provided by residual pathways and temporal modeling, which already mitigate stochastic effects.

\section{Conclusion and future work}
\label{sec:conclusion}

We introduced a Custom Loss Function (CLF) to enhance robustness of the results by mitigating stochastic variability. Experiments on image classification and time series forecasting demonstrated improved performance consistency across diverse architectures. In image classification, CLF significantly reduced accuracy variance, especially in high-variance models. For instance, CLF lowered accuracy standard deviation by 77\% on ShuffleNet-V2 (CIFAR-100). In contrast, stable models like VGG-16 (CIFAR-10) showed smaller gains. Longer CLF activation consistently improved robustness, while late-stage application had limited impact. In time series forecasting, CLF reduced prediction variability, particularly over longer horizons. NLinear saw substantial drops in MAE variability, while more stable models like iTransformer showed modest improvements. CLF was most effective in scenarios with unstable training dynamics or minimal internal regularization. Furthermore, CLF is easily adaptable to other deep learning tasks or architectures, making it suitable for a wide range of experimental setups.

\paragraph{Limitations:}The need for multiple training runs per dataset-architecture pair limits large-scale testing, leaving this area unexplored. Additionally, CLF’s fixed hyperparameters during training may reduce performance, as they do not adapt to changing model dynamics. Dynamically adjusting CLF based on training signals could enhance stability and accuracy further by enabling the model to better respond to varying conditions.

\bibliographystyle{aaai2026}


\end{document}